\def\year{2022}\relax
\documentclass[letterpaper]{article} % DO NOT CHANGE THIS
\usepackage{aaai22}  % DO NOT CHANGE THIS
\usepackage{times}  % DO NOT CHANGE THIS
\usepackage{helvet}  % DO NOT CHANGE THIS
\usepackage{courier}  % DO NOT CHANGE THIS
\usepackage[hyphens]{url}  % DO NOT CHANGE THIS
\usepackage{graphicx} % DO NOT CHANGE THIS
\usepackage{natbib}  % DO NOT CHANGE THIS AND DO NOT ADD ANY OPTIONS TO IT
\usepackage{caption} % DO NOT CHANGE THIS AND DO NOT ADD ANY OPTIONS TO IT
\DeclareCaptionStyle{ruled}{labelfont=normalfont,labelsep=colon,strut=off} % DO NOT CHANGE THIS
\usepackage{algorithm}
\usepackage{algorithmic}
\usepackage[utf8]{inputenc} % allow utf-8 input
\usepackage[T1]{fontenc}    % use 8-bit T1 fonts
\usepackage{hyperref}       % hyperlinks
\usepackage{url}            % simple URL typesetting
\usepackage{booktabs}       % professional-quality tables
\usepackage{amsfonts}       % blackboard math symbols
\usepackage{nicefrac}       % compact symbols for 1/2, etc.
\usepackage{microtype}      % microtypography

\usepackage{cite}
\usepackage{amsmath,amssymb,amsfonts}
\usepackage{algorithmic,algorithm}
\usepackage{graphicx}
\usepackage{textcomp}
\usepackage{xcolor}
\usepackage{subcaption}
\usepackage{threeparttable}
\usepackage{booktabs}

\def\BibTeX{{\rm B\kern-.05em{\sc i\kern-.025em b}\kern-.08em
    T\kern-.1667em\lower.7ex\hbox{E}\kern-.125emX}}
\newcommand{\mat}[1]{\mathbf{#1}}

\newcommand{\graph}{\mathcal{G}}
\newcommand{\nodes}{\mathcal{V}}
\newcommand{\edges}{\mathcal{E}}
\newcommand{\nbhd}[1]{N^{#1}(v)}
\newcommand{\ns}[1]{\overline{N^{#1}(v)}}
\newcommand{\gs}{\overline{\graph}}
\newcommand{\rep}[2]{\mat{h}_{#1}^{#2}}

\newcommand{\supbatch}{\mathcal{B}_s}
\newcommand{\unsupbatch}{\mathcal{B}_u}
\newcommand{\pred}{\hat{\mat{y}}}
\begin{document}

\title{Scalable Consistency Training for Graph Neural Networks \\ via Self-Ensemble Self-Distillation}
\author{Cole Hawkins$^1$\thanks{Work completed while the author was an intern at AWS}, Vassilis N. Ioannidis$^2$, Soji Adeshina$^2$, George Karypis$^2$ \\ {\normalfont $^1$University of California, Santa Barbara \:\:\: $^2$Amazon Web Services}}

\maketitle

\begin{abstract}
Consistency training is a popular method to improve deep learning models in computer vision and natural language processing. Graph neural networks (GNNs) have achieved remarkable performance in a variety of network science learning tasks, but to date no work has studied the effect of consistency training on large-scale graph problems. GNNs scale to large graphs by minibatch training and subsample node neighbors to deal with high degree nodes. We utilize the randomness inherent in the subsampling of neighbors and introduce a novel consistency training method to improve accuracy. For a target node we generate different neighborhood expansions, and distill the knowledge of the average of the predictions to the GNN. Our method approximates the expected prediction of the possible neighborhood samples and practically only requires a few samples. We demonstrate that our training method outperforms standard GNN training in several different settings, and yields the largest gains when label rates are low. 
\end{abstract}

\section{Introduction}

A large body of work improves the predictive accuracy of deep neural networks by promoting consistent internal representations or consistent final predictions of the input data under data augmentation \cite{xie2019unsupervised,grill2020bootstrap,chen2020simple,he2020momentum}. State-of-the-art results on large-scale image datasets, especially in the low-label setting, rely on these techniques to boost performance compared to standard methods that learn representations adhering to a supervised loss \cite{chen2020big,pham2020meta,sohn2020fixmatch}. An open question is whether consistency training methods can provide accuracy improvements on large-scale prediction tasks for graph-structured data, especially when node label rates are low. Early work on supervised graph neural network (GNN) training has focused on small citation datasets \cite{yang2016revisiting,kipf2016semi}. As benchmark GNN datasets grow in size and benchmark prediction tasks grow in complexity \cite{hu2020open} consistency training is a promising approach to improve accuracy.

Recent efforts to transfer consistency training strategies to the graph domain \cite{wang2020nodeaug,feng2020graph} leverage graph data augmentations such as node and edge dropping. Consistency training under thse data augmentations increases accuracy on small graph problems, but existing work relies on strong assumptions that limit transfer to large and complex graph learning tasks. GRAND \cite{feng2020graph} is a method that applies random feature propagation followed by an MLP. This variation of the popular Correct and Smooth framework \cite{huang2020combining} performs well on small datasets but does not extend naturally to scalable training on heterogeneous graphs.  NodeAug \cite{wang2020nodeaug} requires an independent {\it non-overlapping} subgraph expansions which contravenes the assumptions of scalable GNN training in which the overlapping neighborhood expansions of multiple nodes are combined to take advantage of fast kernels for large-scale matrix computation \cite{wang2019deep}.

Our goal is to design a consistency training method that can be applied to large-scale GNN training on graphs at least $1-2$ orders of magnitude larger than those considered in existing work~\cite{feng2020graph,wang2020nodeaug}. Our proposed training method retains the scalability of standard GNN methods, can be used with any GNN model, and does not require assumptions on the minibatch process, or node types.

Scalable GNN training is challenging when the graphs contain a large number of high-degree nodes so common approaches sub-sample node neighbors during the training process \cite{hamilton2017inductive,wang2019deep}. The resulting number of sampled nodes fits in GPU memory and allows minibatch GNN training. However this stochastic neighborhood expansion introduces noise into the GNN inference and training processes. The motivating insight of our method is that the noise induced by neighborhood sampling during large-scale graph training can be used to generate an ensemble of predictions. We call this procedure ``self-ensembling" and demonstrate that it produces results that are competitive with GNN ensembling. Based on this insight, we propose consistency training as a method to take advantage of the self-ensembling accuracy boost during the training process. We incorporated our consistency loss into the GNN training process and demonstrate that it provides accuracy gains of $1\%-2\%$ on node classification on large-scale benchmarks. This paper makes the following contributions.

\noindent{\bf C1.} We present consistency training, a simple and scalable method, to take advantage of self-ensembling and improve test accuracy within a single training and inference run.

\noindent{\bf C2.} We provide a principled motivation for consistency training by casting our method as online distillation of a virtual ensemble.

\noindent{\bf C3.} We provide experiments on both large and small datasets to verify our results. We consider both the transductive and inductive learning settings, and demonstrate that our consistency training method is not only competitive with existing work in the small-data setting, but scales naturally to larger graphs and outperforms standard supervised GNN training, especially in the low-label setting. 

\begin{figure*}
  \begin{subfigure}[t]{0.33\textwidth}
    \includegraphics[width=\textwidth]{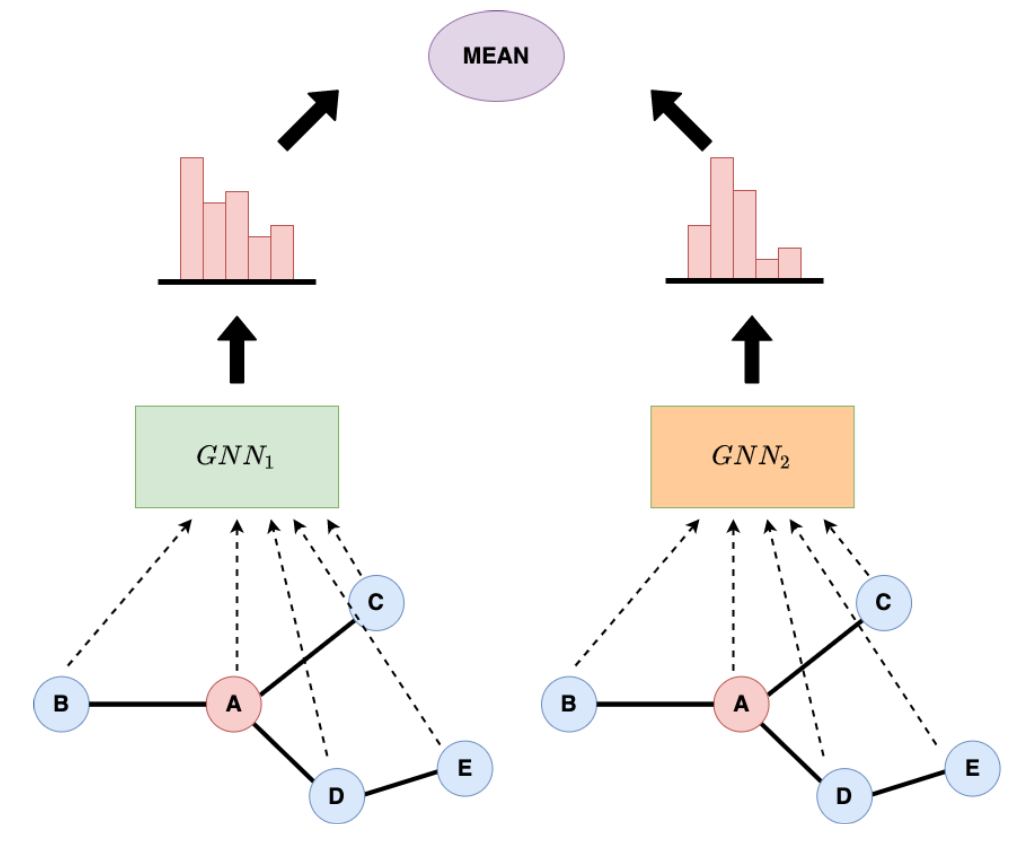}
    \caption{}
    \label{fig: ensemble}
  \end{subfigure}\hfill
  \begin{subfigure}[t]{0.33\textwidth}
    \includegraphics[width=\textwidth]{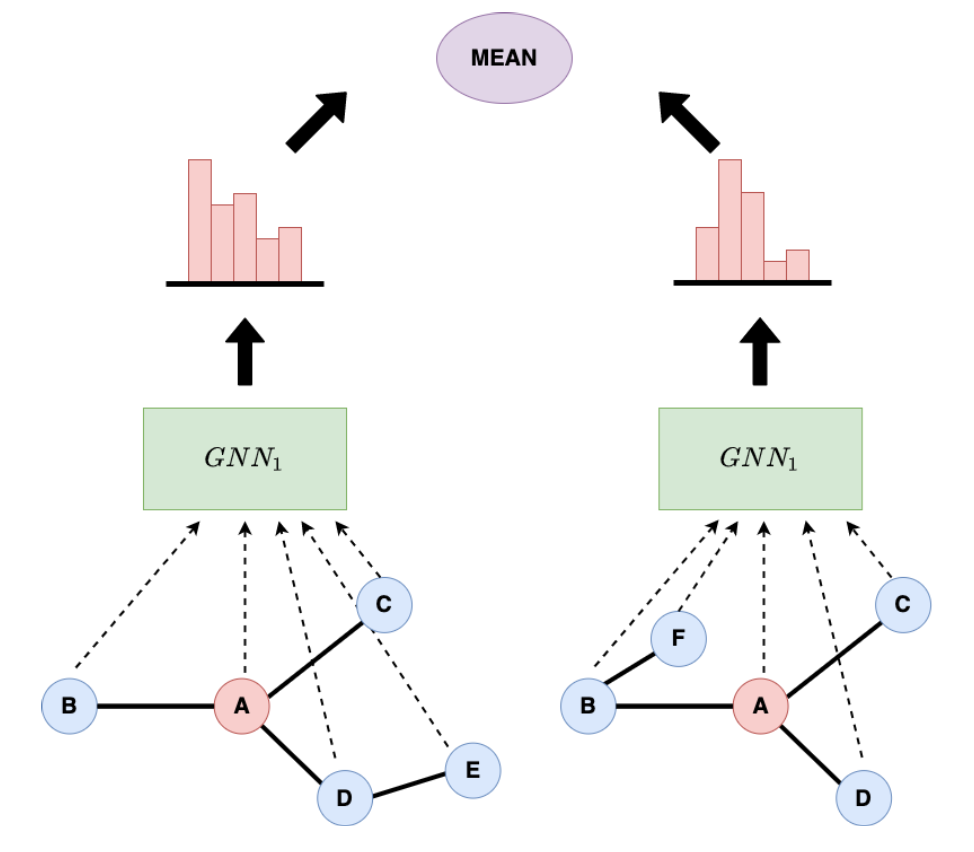}
    \caption{}
    \label{fig: self ensemble}
  \end{subfigure}\hfill
  \begin{subfigure}[t]{0.33\textwidth}
    \includegraphics[width=\textwidth]{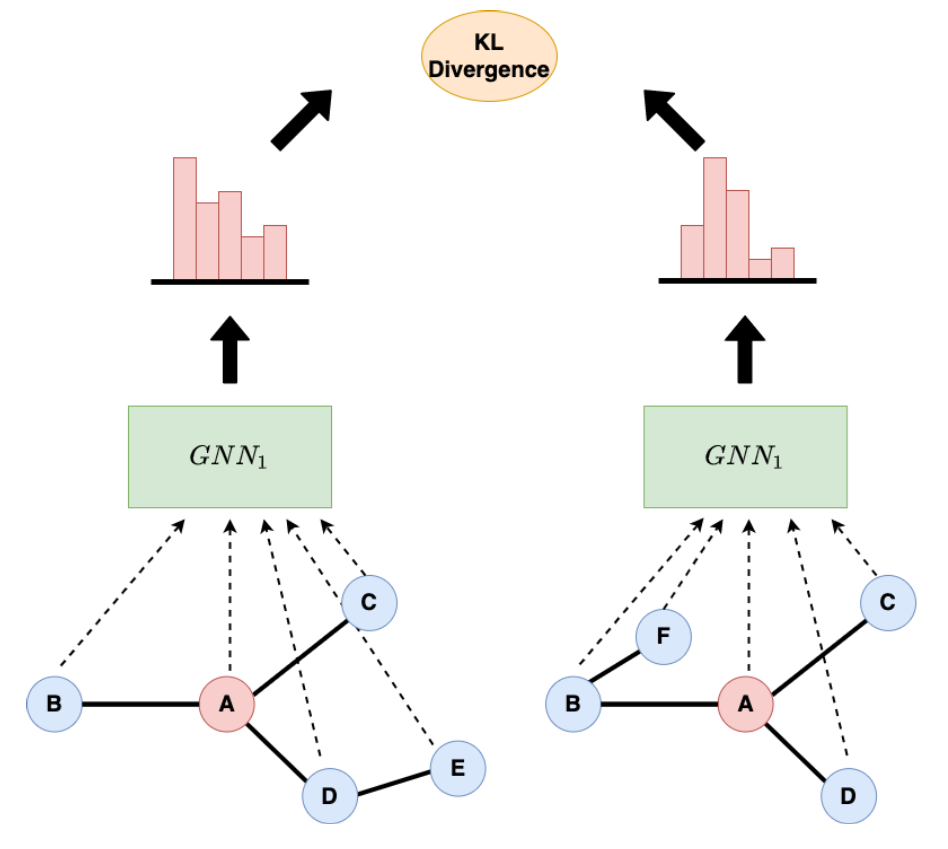}
    \caption{}
    \label{fig: consistency training}
  \end{subfigure}\hfill
\caption{(a) Ensembling multiple models with the same neighborhood expansion. (b) Ensembling multiple neighborhood expansions with the same model. (c) Proposed: single-model self-consistency across multiple neighborhood expansions.}
\label{fig: movtivating fig}
\vspace{-0.15in}
\end{figure*}

\section{Related work}

\subsection{Ensembling and Self-Ensembling}

Ensembling the predictions of multiple models often improves accuracy~\cite{caruana2004ensemble}. Modern neural networks, trained with stochastic gradient descent, can be ensembled to improve accuracy due to the prediction diversity ensured by stochastic training and random weight initializations \cite{lakshminarayanan2016simple}. NARS \cite{yu2020scalable} is a graph learning method that averages over multiple neighborhood representations before outputting a final prediction. NARS differs from our proposed methods in that (1) it does not use GNN predictors and (2) ensembling is performed over internal representations, rather than predictions.

Ensembling is appealing due to its simplicity. However the main drawback is the increase in training and prediction costs, as both scale linearly with the number of models used in an ensemble. First we demonstrate that self-ensembling increases model accuracy at test time. Motivated by this insight, we develop a consistency training procedure to increase model accuracy without requiring multiple predictions.

\subsection{Consistency Training and Self-Supervised Learning}

Both consistency training \cite{xie2019unsupervised} and self-supervised learning \cite{chen2020self,he2020momentum,grill2020bootstrap} improve the accuracy of large-scale computer vision models by enforcing consistency across representations generated under different stochastic augmentations of the data. These approaches are most effective in the low-label, semi-supervised setting. Many works have attempted to transfer these methods to GNN training \cite{feng2020graph,thakoor2021bootstrapped}. Existing work focuses on either the graph property prediction task or small-scale node property prediction tasks \cite{verma2019graphmix,you2020does,wang2020nodeaug}. However efforts to extend these methods to large graph learning problems have underperformed standard supervised approaches \cite{thakoor2021bootstrapped}. An open question is whether GNNs can benefit from self-supervision on large-scale node property prediction problems. The two most relevant works to ours are GRAND \cite{feng2020graph} and NodeAug \cite{wang2020nodeaug}. GRAND uses several rounds of random feature propagation before employing an MLP predictor. This approach relies heavily on the homophily assumption, and is unsuitable for heterogeneous graphs or edge features. NodeAug uses a complex edge add/drop procedure to generate augmentations, but relies on independent perturbations to graph nodes. This requires a complex minibatch generation procedure that is unsuitable for large-scale GNN training. In contrast, our work integrates naturally with scalable GNN training.

\subsection{Distillation and Self-Distillation}

Knowledge distillation is a popular method for transferring the knowledge of a more expensive teacher model to a cheaper student model \cite{hinton2015distilling}. This process is traditionally done in two stages. During the first stage the teacher model is trained to produce high-accuracy predictions. During the second stage the student model is trained to imitate the teacher predictions. Knowledge distillation often leads to higher student accuracy compared to supervised training of the student model alone \cite{hinton2015distilling,cho2019efficacy}. Recent work on self-distillation improves accuracy in computer vision models by using the student model as its own teacher \cite{zhang2019your}. During training the outputs of a model are used as a source of knowledge to supervise the intermediate representations of the same model. Distillation has been applied to graph neural networks (GNNs) in order to reduce prediction latency \cite{yan2020tinygnn,yang2020distilling} and self-distillation has been applied to GNNs to improve prediction accuracy \cite{chen2020self}. Existing GNN self-distillation work relies on a complex series of loss functions to supervise hidden representations \cite{chen2020self}. Our method does not access hidden representations.

%table of notation goes here -use definitions
\section{Notation and Background}

We use lowercase letters $a$ to denote scalars, boldface lowercase letters $\mat{a}$ to denote vectors, and boldface uppercase letters $\mat{A}$ to denote matrices. We represent a graph $\mathcal{G}:=\left(\nodes,\edges\right)$ as a collection of nodes $\nodes:= \{v_1,v_2,\ldots,v_n\}$ and edges $\edges:=\{(v_i,v_j)\in \mathcal{V}\times \mathcal{V}\}$, where $v_i$ is connected to $v_j$. 
In node prediction tasks, each node $v_i$ is associated with a feature vector  $\mat{x}_{{i}} \in \mathbb{R}^{d_0}$, as well as a 
label $\mat{y}_{i}\in\mathbb{R}^{c}$ that is the one-hot vector representation of the $c$ classes. In certain cases we have access to the labels at only a subset of nodes $\{\mat{y}_{i}\}_{{i}\in\mathcal{L}}$, with $\mathcal{L} \subset\nodes$. In this paper we address two learning paradigms, where the task in both is to predict the labels of the unlabeled nodes $\{\mat{y}_{i}\}_{{i}\in\mathcal{U}}$, with $\mathcal{U}=\mathcal{V}\setminus\mathcal{L}$.

\noindent \textbf{Transductive learning}.  Given $\{\mat{y}_{{i}}\}_{{i}\in\mathcal{L}}$,  the feature values of all nodes $\{\mat{x}_{{i}}\}_{{i}\in \mathcal{V}}$, and the connectivity of the graph $\edges$ predict $\{\mat{y}_{i}\}_{{i}\in\mathcal{U}}$.

\noindent\textbf{Inductive learning}.  Given $\{\mat{y}_{{i}}\}_{{i}\in\mathcal{L}}$,  the feature values of the labeled nodes $\{\mat{x}_{{i}}\}_{{i}\in \mathcal{L}}$, and the induced connectivity of the graph by retaining the labeled nodes $\edges_\mathcal{L}:=\{(v_i,v_j): (v_i,v_j) \in \edges \text{ and } v_i,v_j \in \mathcal{L}\}$ predict $\{\mat{y}_{i}\}_{{i}\in\mathcal{U}}$.

To address these tasks we will employ graph neural networks (GNNs). GNNs generate node-level representations via message passing. At layer $l$ the feature representation $\rep{v}{l}$ of node $v\in\nodes$ is determined by aggregating the node representations of its 1-hop neighborhood
\[\nbhd{1}:=\{u|(u,v)\in\edges\}\] 
and then updating the previous node-level representation $\rep{v}{l-1}$. This operation can be expressed as
\begin{equation}
    \label{eq: GNN prop}
        \rep{v}{l} = \text{UPDATE}\left(\rep{v}{l-1},\text{AGGREGATE}\left(\left\{\rep{u}{l-1}|u\in \nbhd{1}\right\}\right)\right).
\end{equation}
The $\text{AGGREGATE}$ and $\text{UPDATE}$ operations are differentiable functions and $\text{AGGREGATE}$ is often required to be permutation-invariant. Given a graph $\graph$, a target node $v$, and a matrix of $d$-dimensional input features $\mat{X}\in \mathbb{R}^{|\nodes|\times d}$, an $l$-layer GNN $g$ produces a prediction $\pred_v:=g\left(v,\graph,\mat{X}\right)$. The prediction is obtained by repeating the message passing operations from~\eqref{eq: GNN prop} $l$ times and then passing the final output through an MLP.

We use the symbol $\gs=(\overline{\nodes},\overline{\edges})$ to represent a subgraph of $\graph$ whose nodes and edges are selected via random subsampling of $\graph=(\nodes,\edges)$. We note that node-dropping and edge-dropping are commonly employed as a data augmentations, and fit within this framework \cite{rong2019dropedge,wang2020nodeaug,verma2019graphmix}. In this paper we limit the candidate stochastic transformations to random node dropping and random edge-dropping and do not add nodes or edges. We follow unsupervised learning literature and use the terms ``views of a datapoint" and ``i.i.d. stochastic augmentations of a datapoint" interchangeably \cite{he2020momentum,chen2020simple}. 

% \begin{table}
%     \centering
%     \caption{Notation}
%     \label{tab:my_label}
%     \begin{tabular}{@{\hspace{2pt}}ll@{\hspace{2pt}}}\toprule
%         Symbol & Description\\ \midrule
%         $a$ & A scalar. \\
%         $\mat{a}$ & A vector. \\
%         $\mat{A}$ & A matrix. \\
%         $\graph$ & A graph. \\
%         $\nodes$ &  The set of nodes in the graph $\graph$.\\
%         $\edges$ &  The set of edges in the graph $\graph$.\\
%         $\mat{X}$ &  The node feature matrix.\\
%         $\nbhd{k}$ &  The $k$-hop neighborhood of a node $v\in\nodes$.\\
%         $\ns{k}$ &  A randomly subsampled $k$-hop neighborhood.\\
%         $g$ & A GNN.\\
%         $\gs$ &  A random perturbation of $\graph$ generated by data augmentation.\\\bottomrule
%     \end{tabular}
% \end{table}

\section{Method: Self-Ensembling and Consistency Training}

    In this section we provide a principled motivation for the practice of consistency training by casting consistency training as distillation from a virtual ensemble. We introduce the notion of self-ensembling, or ensembling across different neighborhood expansions of a node. Next, we demonstrate how self-ensembling naturally results to scalable GNN training and provide both theoretical motivation and empirical justification for consistency training. Finally, we develop our training algorithm from the perspective of self-distillation and present our Self-Ensemble Self-Distillation approach. We also illustrate how our method can be applied to small-scale GNN learning tasks.

\subsection{Neighborhood sampling for scaling GNN Training} 

GNN training and inference on large graphs with high node degrees requires neighborhood sampling \cite{hamilton2017inductive}. 
Instead of using the entire $l$-hop neighborhood expansion $\nbhd{l}$ to perform the forward pass in an $l$-layer GNN, a neighborhood sampler generates a subgraph $\overline{\nbhd{l}}$ of the full $l$-hop neighborhood $\nbhd{l}$. Subgraph sampling is used for GNN message passing during both training and inference.
The challenge presented by this approach is that no subgraph contains full neighborhood information, and this may lead to lower classification accuracy. More importantly to our approach, neighborhood sampling introduces noise in the input which means that the GNN prediction associated with a given node $v$ is non-deterministic. 

\subsection{GNN Self-Ensembling}

Our goal is to take advantage of stochasticity in the model input to improve GNN accuracy. 
From the perspective of ensembling, we exploit the noise to generate predictions which are correct in expectation. 
Instead of looking to model weights for this cheaper source of noise, we turn to the data. 
We overload notation for $\nbhd{k}$ and use $\overline{\nbhd{k}}\sim\nbhd{k}$ to indicate that neighborhood samples (stochastic subgraphs of the $k$-hop ego-network) are generated by subsampling from the neighborhood subgraph $\nbhd{k}$. 
We propose to construct a {\it self-ensemble} by ensembling across different neighborhood subgraph expansions. We construct a prediction 
\begin{equation}
\label{eq: self-ensemble}
\pred = \frac{1}{n}\sum_{i=1}^n g_{\theta}\left(v,\ns{l}_i,\mat{X}\right)
\end{equation}
by averaging over several neighborhood subgraphs. In comparison to full-model retraining, self-ensembling via neighborhood re-sampling is comparatively cheap. 
We demonstrate in Section \ref{sec: experiments self ensembling} that {\it ensembling over different neighborhood subgraph predictions is competitive with ensembling over multiple training runs}. 
Therefore, in a setting with fixed training costs, self-ensembling presents a natural and cheap method to improve accuracy for large-scale GNN inference. 
While self-ensembling addresses the high costs of repeated training runs, this approach does not remove the additional costs at inference time. In our next section, we propose a method designed to take advantage of the benefits of self-ensembling during the training process. Our goal is to achieve the benefits of self-ensembling at inference time by incorporating this idea into the training phase.

\subsection{Consistency Training}

Self-ensembling via neighborhood expansion improves the predictive accuracy at the cost of multiple forward passes. Since self-ensembling is available during the training process, we hope to take advantage of increased accuracy {\it during the training phase}. We hypothesize that the self-ensembled predictions available during training can provide a pseudo-label training target. We propose to add an additional consistency loss to the standard loss function. Given a single training node $v$ we can compute a  pseudo-label $\mat{z}$ and a corresponding loss function $l$ as follows:\\

\noindent 1. Given node $v$ compute $n$ $k$-layer neighborhood expansions
\begin{equation}
    \label{eq: consistency preds}
    \pred_i =  g(v,\ns{l}_i,\mat{X}) \text{ for }i=1\dots n, \ns{l}\sim \nbhd{l}.
\end{equation}
2. Compute the mean prediction:
\begin{equation}
    \label{eq: self-ensemble average}
\mat{z} = \frac{1}{n}\sum_{i=1}^n \pred_i
\end{equation}
3. Use the mean prediction $\mat{z}$ to compute the loss for node $v$:
\begin{equation}
\label{eq: unsharpened consistency loss}
l(v) = \frac{1}{n}\sum_{i=1}^n KL(\pred_i || \mat{z})
\end{equation}

This loss function can be applied to all nodes during the training process, including nodes that are not labeled. This is particularly advantageous in the large-scale transductive setting when few labels are available. In such cases, we demonstrate experimentally that an additional source of supervision can significantly increase accuracy. The loss function in Equation \ref{eq: unsharpened consistency loss} is suitable for our goal on the labeled nodes, but admits a pathological solution on the unlabeled nodes. This trivial solution is achieved by assigning $\{\pred_i\}$ to be the highest entropy solution which places equal probability in all classes. To prevent this trivial solution we apply temperature sharpening to the mean prediction and set the final consistency loss as 
\begin{equation}
\label{eq: consistency loss}
l_{con}(v) = \frac{1}{n}\sum_{i=1}^n KL(\pred_i || \mat{z}^{1/T_{con}}).
\end{equation}
We avoid oversmoothing of the model outputs by setting the consistency sharpening temperature $T_{con}\in (0,1)$ In practice we have found that $T_{con}=0.4$ is a strong default and $n=2$ neighborhood expansions is sufficient to capture the benefits of self-ensembling. We provide a parameter sensitivity study in Section \ref{sec: parameter sensitivity}. During training we use a weighted combination of the supervised loss $l_{sup}$ and the consistency loss $l_{con}$. Let $\supbatch,\unsupbatch$ be a batch of labeled nodes and a batch of nodes that are not necessarily labeled (since the consistency loss is applied to all nodes). The full batch consistency loss is computed as follows: 
\begin{equation}
    \label{eq: full consistency loss}
    l(\mathcal{B}_s,\mathcal{B}_u) = \frac{1}{|\supbatch|}\sum_{v\in\supbatch} l_{sup}(v)+ \frac{\alpha}{|\unsupbatch|}\sum_{v\in\unsupbatch} l_{con}(v)
\end{equation}
where $\alpha$ is a user-specified hyperparameter. We follow \cite{xie2019unsupervised} and mask the loss signal of the labeled nodes on which the model is highly confident early in the learning process. This prevents the model from overfitting the supervised objective early in the learning process. The masking threshold is linearly increased from $1/c$ to 1 over the course of the learning process where $c$ is the class size.%Our full algorithm is presented in Algorithm \ref{alg: consistency training}.
% \begin{algorithm}
% \label{alg: consistency training}
% \begin{algorithmic}
% \REQUIRE Inputs
% \STATE Consistency training algorithm goes here.
% \RETURN Outputs
% \end{algorithmic}
% \end{algorithm}

\subsection{Consistency Training as Self-Ensemble Self-Distillation \label{sec: self ensemble self-distillation}}

Distillation is a popular technique to improve the predictions of neural networks \cite{hinton2015distilling}. The most common procedure is to train a teacher model $g_{t}$ first. Then a student model $g_{s}$ is trained using a weighted combination of the standard supervised loss $l_{sup}$ and a distillation loss $l_{dis}$:
\begin{equation}
    \label{eq: distillation loss}
    \begin{split}
    l_{dis}(v) &= KL\left(g_s(v)^{1/T_{dis}}||\:g_t(v)^{1/T_{dis}}\right)\\
    l(v) &= l_{sup}\left(v\right)+\alpha l_{dis}(v).
    \end{split}
\end{equation}
The distillation loss $l_{dis}$ forces the predictions of the student to be similar to those of the teacher and $l_{sup}$ is the standard supervised loss. The distillation temperature $T_{dis}$ can be used to smooth model predictions for better transfer \cite{hinton2015distilling,cho2019efficacy}. 

We re-interpret our consistency training framework as online distillation from a self-ensemble. 
Let $g_{\theta}$ be a single model. We define the self-ensemble teacher $g_t$ by
\begin{equation}
    g_t(v) := \left(\frac{1}{n}\sum_{i=1}^n g_{\theta}(v,\ns{l}_i,\mat{X})\right)^{1/T_{con}}
\end{equation}
where $T_{con}$ refers to the sharpening temperature for consistency training. Next we use the self-ensemble model $g_t$ as the teacher in Equation \eqref{eq: distillation loss}. We set the student model as $g_s = g_\theta$ and select the distillation temperature $T_{dis}=1$ to arrive at exactly the consistency loss from Equation \eqref{eq: consistency loss}. Different from standard neural network distillation, our method is applied in an online manner as both the student and teacher improve during training.

\section{Experimental Results \label{sec: experiments}}
\begin{table*}[htbp]
\caption{Dataset Characteristics}
\begin{center}\scalebox{.8}{ 
\begin{tabular}{|c|r|r|r|r|r|}
\hline
\textbf{Name}& \textbf{Nodes} & \textbf{Edges} & \textbf{Features} & \textbf{Classes} & \textbf{Train/Val/Test} \\ \hline
ogbn-arxiv & 169,323 & 1,166,243 & 128 & 40 & 55 / 17 / 28 \\ \hline
ogbn-products & 2,449,029 & 61,859,140 & 100 & 47 & 8 / 2 / 90 \\ \hline
reddit & 232,965 & 11,606,919 & 602 & 41 &  66 / 10 / 24 \\ \hline
\end{tabular}}
\label{tab: datasets}
\end{center}
\end{table*}

We test all three methods discusses in this paper (ensembling, self-ensembling, and consistency training) on a range of node classification tasks for graph datasets of widely varing sizes (see Table \ref{tab: datasets}). The Reddit dataset uses the standard split and is obtained from the Deep Graph Library (DGL) \cite{wang2019deep}. The ogbn-arxiv and ogbn-products datasets use the standard Open Graph Benchmark split \cite{hu2020open}. We used the ogbn-arxiv dataset as a test case to evaluate scalable training approaches, and used neighborhood sampling for training and inference. It is possible to train ogbn-arxiv with full-batch training and full-graph inference to improve accuracy results. We use a graph convolutional network (GCN) \cite{kipf2016semi} and a graph attention network (GAT) \cite{velivckovic2017graph} as our baseline models. For ogbn-arxiv, ogbn-products, and Reddit inference and training is performed using neighborhood fanouts of 5,5, and 20 respectively. All methods are implemented using DGL \cite{wang2019deep} and PyTorch \cite{paszke2019pytorch}. We report all results as $\mu\pm\sigma$ where the mean $\mu$ and standard deviation $\sigma$ are computed over 10 runs for ogbn-arxiv and 5 runs for Reddit/ogbn-products. Results on small benchmark datasets (Cora/Citeseer/Pubmed) are presented in the appendix. Results are obtained using an AWS P3.16xlarge instance with 8 NVIDIA Tesla V100 GPUs.

\subsection{Multiple Models vs Multiple Views\label{sec: experiments self ensembling}}

First, we compare the benefits of ensembling vs self-ensembling on the ogbn-arxiv and reddit datasets for both the GAT and GCN models to empirically motivate our consistency training method. We focus on the transductive setting for this comparison. On the ogbn-arxiv dataset we test a GCN model (Table \ref{tab: arxiv gcn transductive self-ensemble}). We observe that self-ensembling not only improves accuracy, but is actually competitive with model ensembling accuracy at inference time. We find similar results for a GAT model on ogbn-arxiv as well as both GCN and GAT models on reddit in the appendix.

\begin{table*}[htbp]
\caption{ogbn-arxiv test accuracy, Ensemble vs Self-Ensembling, Transductive Setting, GCN-3 Fanout-5}
\begin{center}\scalebox{.8}{ 
\begin{tabular}{|c|c|c|c|c|c|c}
\hline
\textbf{Number}&\multicolumn{5}{|c|}{\textbf{Number of Models}} \\
\cline{2-6}\textbf{of Views} & 1 & 2 & 3 & 4 & 5 \\
\hline
1 & 70.23$\pm$0.27 & 70.38$\pm$0.32 & 70.70$\pm$0.27 & 70.71$\pm$0.22 & 70.81$\pm$0.13  \\\hline
2 & 70.78$\pm$0.26 & 71.03$\pm$0.32 & 71.10$\pm$0.24 & 71.29$\pm$0.17 & 71.26$\pm$0.11  \\\hline
3 & 71.03$\pm$0.29 & 71.22$\pm$0.30 & 71.18$\pm$0.40 & 71.39$\pm$0.27 & 71.44$\pm$0.15  \\\hline
4 & 70.93$\pm$0.39 & 71.37$\pm$0.34 & 71.54$\pm$0.20 & 71.45$\pm$0.25 & 71.38$\pm$0.17  \\\hline
5 & 71.02$\pm$0.33 & 71.49$\pm$0.17 & 71.37$\pm$0.31 & 71.51$\pm$0.21 & 71.58$\pm$0.09  \\\hline
\end{tabular}}
\label{tab: arxiv gcn transductive self-ensemble}
\end{center}
\end{table*}

\subsection{Consistency Training Parameter Selection \label{sec: parameter sensitivity}}

\begin{figure*}
  \begin{subfigure}[t]{0.33\textwidth}
    \includegraphics[width=\textwidth]{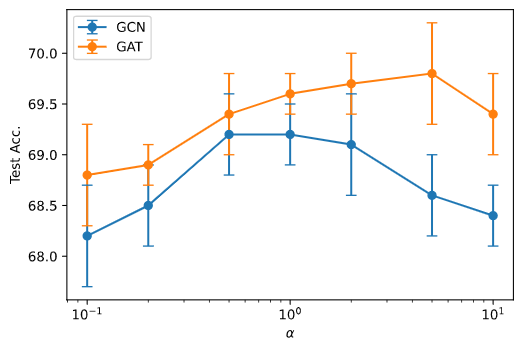}
    \caption{}
    \label{fig: alpha sensitivity}
  \end{subfigure}\hfill
  \begin{subfigure}[t]{0.33\textwidth}
    \includegraphics[width=\textwidth]{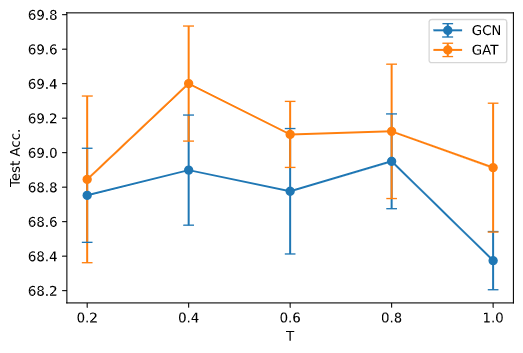}
    \caption{}
    \label{fig: temp sensitivity}
  \end{subfigure}\hfill
  \begin{subfigure}[t]{0.33\textwidth}
    \includegraphics[width=\textwidth]{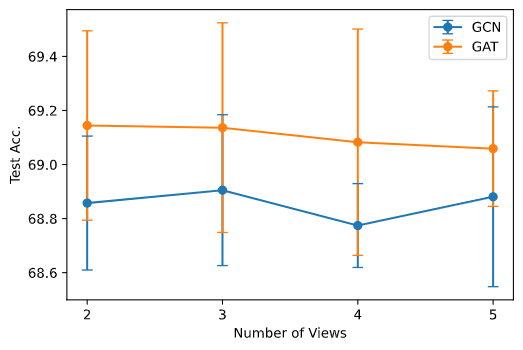}
    \caption{}
    \label{fig: views sensitivity}
  \end{subfigure}\hfill
\caption{Parameter sensitivity studies for GCN-2 and GAT-2 models on ogbn-arxiv. Sensitivities to (a) consistency loss weight $\alpha$, (b) sharpening temperature $T$, and (c) number of self-ensembled views $n$. }
\label{fig: sensitivity studies}
\vspace{-0.1in}
\end{figure*}

Before we present our results on consistency training, we present a parameter sensitivity study. Hyperparameter selection is described in the Appendix. There are three model parameters that define our consistency training model: (1) number of views $n$, (2) weighting parameter $\alpha$, and (3) sharpening temperature $T_{con}$. First we fix $n=2$ and $T=0.4$ to study the effects of varying $\alpha$. The setting $T=0.4$ mirrors prior work in unsupervised learning for image classification models \cite{berthelot2019mixmatch,xie2019unsupervised}. Figure \ref{fig: alpha sensitivity} considers the performance of our models under different settings of $\alpha$. The best $\alpha$ value for the GAT model is $5.0$ and the best value for the GCN model is $1.0$. In general we found that $0.1-5.0$ is an effective range for $\alpha$ across models. We fix $\alpha=2.0$ as compromise between the two models for our other parameter studies in this section. We continue with $n=2,\alpha=2.0$ and study the sensitivity of consistency training to the sharpening temperature parameter $T_{con}$. We observe from Figure \ref{fig: temp sensitivity} that the GAT model performs best with temperature $T=0.4$, while the GCN model has comparable performance for $T\in\{0.2,0.4,0.6,0.8\}$. Since $T=0.4$ is also an acceptable setting for the GCN model, yielding the second best result, we fix this this value for all our experiments. Finally we observe from Figure \ref{fig: views sensitivity} that increasing the number of views yields little to no benefit for either model. 

For the ogbn-arixv/Reddit/ogbn-products datasets we fix $T_{con}=0.4,n=2$, and sweep across the grid $\alpha\in\{0.05,0.1,0.2,0.5,1.0,2.0,5.0\}$ with two trials per setting to select $\alpha$ based on validation set accuracy, and then run the pre-specified number of trials to obtain our mean and standard deviation. We discuss parameter tuning for the small datasets in the Appendix.

\subsection{Consistency Training vs Single-View Single-Model}

\begin{figure*}
  \begin{subfigure}[t]{0.4\textwidth}
    \includegraphics[width=\textwidth]{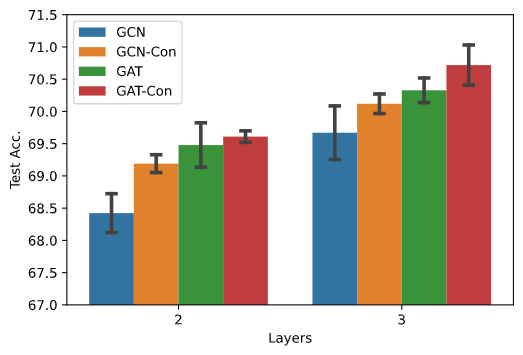}
    \caption{High label rate training.}
    \label{fig: arxiv high label}
  \end{subfigure}\hfill
  \begin{subfigure}[t]{0.4\textwidth}
    \includegraphics[width=\textwidth]{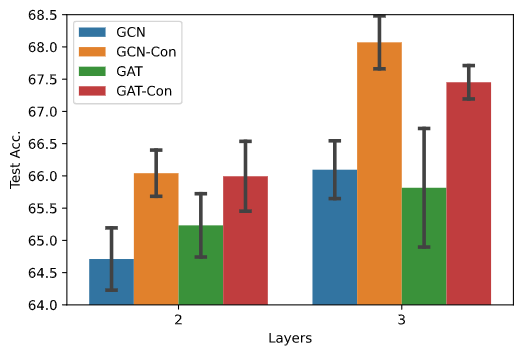}
    \caption{Low label rate training.}
    \label{fig: arxiv low label}
  \end{subfigure}\hfill
\caption{Inductive results on ogbn-arxiv using fanout-5 models for a range of models and label rates. (a) Results with $55\%$ of labels used during training. (b) Results with $5.5\%$ of labels used for training. ``-Con" denotes consistency training. Error bars are one standard deviation.}
\label{fig: arxiv consistency results multiple models}
\vspace{-0.2in}
\end{figure*}

\begin{figure*}
  \begin{subfigure}[t]{0.4\textwidth}
    \includegraphics[width=\textwidth]{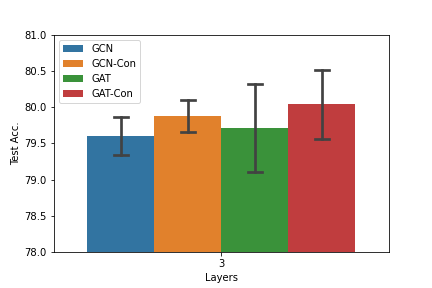}
    \caption{High label rate training.}
    \label{fig: products high label}
  \end{subfigure}\hfill
  \begin{subfigure}[t]{0.4\textwidth}
    \includegraphics[width=\textwidth]{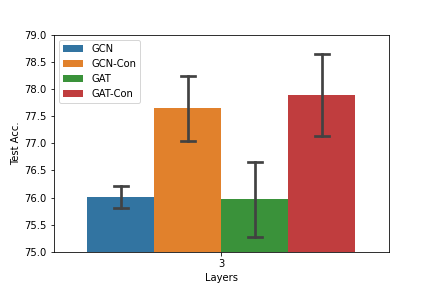}
    \caption{Low label rate training.}
    \label{fig: products low label}
  \end{subfigure}\hfill
\caption{Transductive results for fanout-5 GAT/GCN models on ogbn-products for a range of label rates. (a) Results with $8.0\%$ of nodes in the training set. (b) Results with $0.8\%$ of nodes in the training set.}
\label{fig: products consistency results multiple models}
\vspace{-0.2in}
\end{figure*}

\begin{figure}
\centering
    \includegraphics[width=0.35\textwidth]{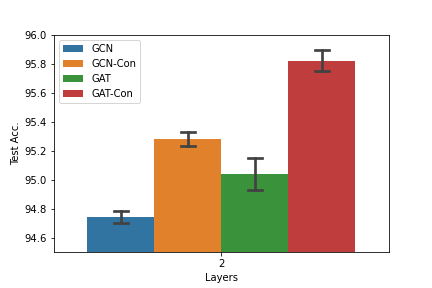}
\caption{Inductive results on Reddit using fanout-20 models for a range of models with $6.6\%$ of labels used for training. }
\label{fig: reddit consistency results low label}
\vspace{-0.2in}
\end{figure}

In this section we demonstrate the benefits of consistency training. For each dataset we test the performance of our models on high and low label rates. In the high label rate setting we use all available training labels and in the low label rate setting we use $10\%$ of available training labels. First, we report the results on ogbn-arxiv on a wide variety of architectures and label rates in Figure \ref{fig: arxiv consistency results multiple models}. We consider the inductive setting for three layer and two layer GCN and GAT models. We also test two label rate settings, $55\%$ and $5.5\%$. The high label rate of $55\%$ is the standard benchmark ogbn-arxiv label setting and the low label rate of $5.5\%$ uses only $10\%$ of the available training labels. We randomly subsample the $10\%$ from the existing training labels for each low label rate run. We observe that consistency training improves results for all combinations of models and label rates. The gains are highest when label rates are low. Next, we report results on Reddit, our second-largest dataset in Figure \ref{fig: reddit consistency results low label}. We observe from this experiment that consistency training improves performance most in the low-label setting, but provides modest gains for the GAT model in the full-label setting. Finally, we report results on our largest dataset, ogbn-products, in Figure \ref{fig: products consistency results multiple models}. The default split for ogbn-products only includes the most popular $8\%/2\%$ of products (nodes) for training and validation respectively.  This does not lead to a realistic inductive prediction task so we use the transductive setting for comparison. Again, our consistency training method performs best in the lowest label setting, improving both the GCN and GAT models. In the full-label setting (still at only $8\%$ label rate) consistency training does not provide a significant boost in accuracy. 


\subsection{Consistency Training and Self-Ensembling \label{sec: consistency+self ensemble}}
\begin{table*}[htbp]
\caption{ogbn-arxiv test accuracy, Ensemble vs Self-Ensembling, Inductive Setting, GAT-3 Fanout-5.\\ Results presented as without/with consistency training.}
\begin{center}
\scalebox{.7}{  
\begin{tabular}{|c|r|r|r|r|r|r}
\hline
\textbf{Number}&\multicolumn{5}{|c|}{\textbf{Number of Models}} \\
\cline{2-6}\textbf{of Views} & 1 & 2 & 3 & 4 & 5 \\
\hline
1 & 70.28$\pm$0.27 / 70.63$\pm$0.13 & 70.58$\pm$0.26 / 70.94$\pm$0.13 & 70.72$\pm$0.21 / 70.99$\pm$0.24 & 70.80$\pm$0.17 / 71.05$\pm$0.16 & 70.95$\pm$0.15 / 71.04$\pm$0.14 \\\hline
2 & 70.61$\pm$0.19 / 70.91$\pm$0.26 & 71.04$\pm$0.18 / 71.19$\pm$0.22 & 71.12$\pm$0.20 / 71.31$\pm$0.23 & 71.21$\pm$0.18 / 71.24$\pm$0.15 & 71.15$\pm$0.14 / 71.37$\pm$0.17 \\\hline
3 & 70.84$\pm$0.21 / 70.97$\pm$0.15 & 70.96$\pm$0.23 / 71.35$\pm$0.18 & 71.24$\pm$0.16 / 71.51$\pm$0.27 & 71.20$\pm$0.24 / 71.43$\pm$0.18 & 71.32$\pm$0.18 / 71.48$\pm$0.16 \\\hline
4 & 70.77$\pm$0.30 / 71.15$\pm$0.33 & 71.12$\pm$0.31 / 71.33$\pm$0.12 & 71.22$\pm$0.19 / 71.58$\pm$0.17 & 71.31$\pm$0.14 / 71.47$\pm$0.13 & 71.45$\pm$0.13 / 71.60$\pm$0.14 \\\hline
5 & 70.82$\pm$0.26 / 71.39$\pm$0.38 & 71.27$\pm$0.17 / 71.37$\pm$0.24 & 71.30$\pm$0.07 / 71.53$\pm$0.18 & 71.47$\pm$0.11 / 71.51$\pm$0.09 & 71.40$\pm$0.11 / 71.57$\pm$0.18 \\\hline
\end{tabular}}
\label{tab: arxiv gat inductive self-ensemble}
\end{center}
\end{table*}

\begin{table*}[htbp]
\caption{ogbn-arxiv test accuracy, Ensemble vs Self-Ensembling, Inductive Setting, GCN-3 Fanout-5.\\ Results presented as without/with consistency training.}
\begin{center}
\scalebox{.7}{ 
\begin{tabular}{|c|r|r|r|r|r|r}
\hline
\textbf{Number}&\multicolumn{5}{|c|}{\textbf{Number of Models}} \\
\cline{2-6}\textbf{of Views} & 1 & 2 & 3 & 4 & 5 \\
\hline
1 & 69.67$\pm$0.60 / 70.12$\pm$0.15 & 70.22$\pm$0.24 / 70.40$\pm$0.11 & 70.31$\pm$0.28 / 70.44$\pm$0.13 & 70.30$\pm$0.24 / 70.45$\pm$0.07 & 70.55$\pm$0.28 / 70.49$\pm$0.10 \\\hline
2 & 70.21$\pm$0.59 / 70.68$\pm$0.11 & 70.68$\pm$0.27 / 70.79$\pm$0.18 & 70.94$\pm$0.12 / 70.92$\pm$0.10 & 71.00$\pm$0.25 / 70.95$\pm$0.05 & 70.91$\pm$0.17 / 70.97$\pm$0.13 \\\hline
3 & 70.40$\pm$0.62 / 70.82$\pm$0.16 & 70.86$\pm$0.36 / 71.05$\pm$0.09 & 71.11$\pm$0.26 / 71.09$\pm$0.07 & 70.91$\pm$0.26 / 71.15$\pm$0.13 & 71.09$\pm$0.25 / 71.14$\pm$0.06 \\\hline
4 & 70.60$\pm$0.55 / 70.91$\pm$0.14 & 70.91$\pm$0.47 / 71.08$\pm$0.10 & 71.00$\pm$0.23 / 71.17$\pm$0.08 & 71.13$\pm$0.27 / 71.22$\pm$0.09 & 71.26$\pm$0.24 / 71.24$\pm$0.06 \\\hline
5 & 70.79$\pm$0.47 / 70.96$\pm$0.20 & 71.04$\pm$0.32 / 71.16$\pm$0.06 & 71.23$\pm$0.20 / 71.26$\pm$0.09 & 71.24$\pm$0.22 / 71.25$\pm$0.07 & 71.20$\pm$0.25 / 71.26$\pm$0.08 \\\hline
\end{tabular}}
\label{tab: arxiv gcn inductive self-ensemble}
\end{center}
\end{table*}

Next, we study the interaction between consistency training and self-ensembling in the inductive setting. Our reasoning from Section \ref{sec: self ensemble self-distillation} leads us to believe that single-view consistency-trained models should be competitive with multi-view baseline models. In Table \ref{tab: arxiv gat inductive self-ensemble} we test this hypothesis by comparing ensembled/self-ensembled GAT-3 models trained without/with consistency training on ogbn-arxiv. The results in Table \ref{tab: arxiv gat inductive self-ensemble} demonstrate that consistency training achieves its goal of self-ensemble self-distillation. We can observe from the first column of Table \ref{tab: arxiv gat inductive self-ensemble} that the two-view self-ensembled standard model and the single-view consistency model achieve almost exactly the same accuracy (70.61 and 70.63). Additional results in Table \ref{tab: arxiv gcn inductive self-ensemble} shows similar outcomes for the GCN model trained with consistency loss. The consistency model using $n$ views can even outperform a standard model using $n+1$ views. In this subsection we focused on the inductive setting to demonstrate that when latency constraints exist, a model trained with consistency loss can outperform a model that requires $2\times$ as many forward passes.

\section{Which Nodes are Improved?}
\begin{figure}
    \centering
    \includegraphics[width=0.7\linewidth]{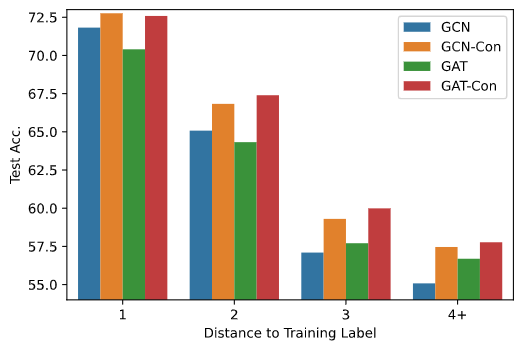}
    \caption{Comparison of ogbn-arxiv three layer model accuracy based on target node distance to a training label with label rate $5.5\%$. ``Con" denotes consistency training.}
    \label{fig: label distance}
\end{figure}

In this section we investigate which node predictions are improved by our consistency training model. In Figure \ref{fig: label distance} we plot the test accuracy (y-axis) and the length of the shortest path from the target node to a node that has a training label (x-axis) to study which nodes in the test set have their predictions improved by the consistency training procedure. Figure \ref{fig: label distance} shows that the benefits of consistency training are distributed relatively evenly for both model types. This differs from fully unsupervised approaches on small graphs \cite{chen2020distance} in which accuracy benefits are concentrated on nodes with high shortest path distance to a training label.

\section{Discussion and Conclusion}

In this paper we presented three methods to improve accuracy of node prediction tasks for graph neural networks: ensembling, self-ensembling, and consistency training. We demonstrated that self-ensembling, or ensembling across multiple neighborhood fanouts, produces predictions competitive with model ensembling. Based on this insight we introduced consistency training as a form of self-ensemble self-distillation during the training phase. We provided experiments demonstrating that consistency training enables single-view single models can capture the benefits of self-ensembling in a single forward pass. The two sources of novelty in this work are (1) self-ensembling and (2) scalable consistency training for GNNs. In (1) we identified a cheap source of increased accuracy at test time and demonstrated its performance across several datasets. Our consistency training method (2) provides a simple and scalable approach to integrate consistency training approaches with GNN training. Future work will investigate the benefits of consistency training in heterogeneous graph datasets and alternate tasks, i.e. link prediction.

%With Neptune ML, you can improve the accuracy of most predictions for graphs by over 50% when compared to making predictions using non-graph methods.
%Making accurate predictions on graphs with billions of relationships can be difficult and time consuming. Existing ML approaches such as XGBoost can’t operate effectively on graphs because they are designed for tabular data. As a result, using these methods on graphs can take time, require specialized skills from developers, and produce sub-optimal predictions.
%Using the Deep Graph Library (DGL), an open-source library to which AWS contributes, that makes it easy to apply deep learning to graph data, Neptune ML automates the heavy lifting of selecting and training the best ML model for graph data, and lets users run machine learning on their graph directly using Neptune APIs and queries. As a result, you can now create, train, and apply ML on Amazon Neptune data in hours instead of weeks without the need to learn new tools and ML technologies.

\bibliography{references.bib}
\end{document}

\title{Appendix: Scalable Consistency Training for Graph Neural Networks via Self-Ensemble Self-Distillation}

\clearpage
\newpage

\appendix

\section{Consistency Training on Small Graphs}

While our motivation is to take advantage of neighborhood fanout, this setting does not apply to small graph GNN training. In the small graph setting the entire graph can be placed into GPU memory, and no neighborhood expansion is necessary. Instead of a neighborhood sampling-based self-ensembling procedure we generate consistency loss predictions by subsampling the entire graph (random node dropping). We generate predictions by
\begin{equation}
    \label{eq: consistency preds full graph}
    \pred_i =  g(v,\gs_i,\mat{X}) \text{ for }i=1\dots n, \gs_i\sim \graph.
\end{equation}
The only change is that we sample the entire graph $\gs_i$ rather than a neighborhood fanout $\ns{l}_i$. Despite this fact, we demonstrate in Section \ref{sec: small graph experiments} that consistency training still provides an advantage on small graph GNN benchmarks. We hypothesize that the accuracy improvement we generate on small graphs (see Section \ref{sec: small graph experiments}) is due to a regularization effect, rather than an ensembling benefit.

\section{Consistency Training Results on Small Graphs \label{sec: small graph experiments}}

Training and inference on small graph datasets is performed using the full graph, so self-ensembling across neighborhood expansions is not a useful approach to increase accuracy. In this setting our proposed consistency training method can still be applied, but we use a different source of stochasticity. We employ node dropping as a method of generating graph samples. We describe the datasets on which we test our method in Table \ref{tab: small dataset characteristics}. Our results on the three small citation datasets are shown in Table \ref{tab: small datasets}. For NodeAug and GRAND we report results from the previously published work. We select the regularization parameter $\alpha\in\{1.0,5.0,10.0,15.0\}$ the node dropout rate from $\{0.05,0.1,0.2\}$, the GCN hidden unit size from $\{16,32,64\}$, and the number of GAT heads from $\{8,16\}$.  We run each parameter setting 10 times, select the configuration with the best validation accuracy, and run that setting for 100 repeats to generate our results. We re-implemented the GCN and GAT baselines to use as  the backbone for our proposed method and took the reported results from GRAND and NodeAug. All means and deviations are reported over 100 runs.  

Applying our method in the small dataset context with node dropping is equivalent to NodeAug with a simplified data augmentation scheme. NodeAug takes advantage of complex and independent data augmentations that are not practical in large-graph training. However the more advanced data augmentations employed by NodeAug outperform our method when the graph size is small.
\begin{table*}[htbp]
\caption{Small Dataset Characteristics}
\begin{center}
\begin{tabular}{|c|r|r|r|r|r|}
\hline
\textbf{Name}& \textbf{Nodes} & \textbf{Edges} & \textbf{Features} & \textbf{Classes} & \textbf{Train/Val/Test} \\ \hline
cora & 2,708 & 10,556 & 1,433 & 7 &  5 / 18 / 37 \\ \hline
citeseer & 3,327 & 9,228 & 3,703 & 6 &  3 / 15 / 30 \\ \hline
pubmed & 19,717 & 88,651 & 500 & 3 &   0.3 / 3 / 5 \\ \hline
\end{tabular}
\label{tab: small dataset characteristics}
\end{center}
\end{table*}
\begin{table}[htbp]
\caption{Small Dataset Results}
\begin{center}
\begin{tabular}{|c|c|r|r|r|}
\hline
Model & Method &Cora & Citeseer & Pubmed \\\hline
GCN & baseline & 81.0$\pm$0.7 & 70.1$\pm$0.8 & 79.3$\pm$0.5 \\\hline
GCN & GRAND \cite{feng2020graph} & 84.5$\pm$0.3 & 74.2$\pm$0.3 & 80.0$\pm$0.3 \\\hline
GCN & NodeAug \cite{wang2020nodeaug} & 84.3$\pm$0.5 & 74.9$\pm$0.5 & 81.5$\pm$0.5 \\\hline
GCN & Ours & 82.7$\pm$0.5 & 75.4$\pm$0.2 & 80.4$\pm$0.1 \\\hline
GAT & baseline  & 82.1$\pm$0.7 & 71.2$\pm$0.6 & 78.6$\pm$0.3\\\hline
GAT & GRAND \cite{feng2020graph} &84.3$\pm$0.4 & 73.2$\pm$0.4 & 79.2$\pm$0.6   \\\hline
GAT & NodeAug\cite{wang2020nodeaug} & 84.8$\pm$0.2 & 75.1$\pm$0.5 & 81.6$\pm$0.6  \\\hline
GAT & Ours & 82.8$\pm$0.5 & 74.3$\pm$0.5 & 79.0$\pm$0.4  \\\hline
%GAT-CON & & &  \\\hline
\end{tabular}
\label{tab: small datasets}
\end{center}
\end{table}

\section{Additional Results: GNN Ensembling+Self-Ensembling \label{appendix: additional se}}

If our two sources of noise (model training re-runs and neighborhood sampling) are uncorrelated, then we expect combining both will yield increase benefits.
We can average predictions over both multiple neighborhood expansions and multiple ensemble models, leading to the final prediction 
\begin{equation}
\label{eq: ensemble + self-ensemble}
\pred = \frac{1}{n_1n_2}\sum_{i=1}^{n_1}\sum_{j=1}^{n_2} g_{\theta_i}\left(v,\ns{l}_j,\mat{X}\right)
\end{equation}
This approach incurs both the benefits and expenses of self-ensembling and ensembling. This approach is expensive during both the training and inference stages. However, as we demonstrate, this approach yields the highest accuracy. We provide additional results in this section on GAT models to demonstrate the effect of self-ensembling (Tables \ref{tab: arxiv gat transductive self-ensemble} and \ref{tab: reddit gat transductive se}). We also provide results for the GCN model on the reddit dataset in Table \ref{tab: reddit gcn transductive se}.

\begin{table*}[htbp]
\caption{ogbn-arxiv test accuracy, Ensemble vs Self-Ensembling, Transductive Setting, GAT-3 Fanout 5}
\begin{center}\scalebox{.9}{ 
\begin{tabular}{|c|c|c|c|c|c|c}
\hline
\textbf{Number}&\multicolumn{5}{|c|}{\textbf{Number of Models}} \\
\cline{2-6}\textbf{of Views} & 1 & 2 & 3 & 4 & 5 \\
\hline
1 & 70.31$\pm$0.45 & 70.68$\pm$0.45 & 70.88$\pm$0.22 & 70.99$\pm$0.14 & 70.82$\pm$0.24 \\\hline
2 & 70.57$\pm$0.38 & 71.18$\pm$0.23 & 71.16$\pm$0.19 & 71.36$\pm$0.25 & 71.25$\pm$0.28  \\\hline
3 & 70.64$\pm$0.45 & 71.13$\pm$0.21 & 71.38$\pm$0.24 & 71.44$\pm$0.18 & 71.40$\pm$0.14 \\\hline
4 & 70.67$\pm$0.42 & 71.18$\pm$0.35 & 71.48$\pm$0.19 & 71.37$\pm$0.24 & 71.31$\pm$0.36 \\\hline
5 & 70.72$\pm$0.44 & 71.37$\pm$0.31 & 71.43$\pm$0.20 & 71.57$\pm$0.17 & 71.59$\pm$0.12 \\\hline
\end{tabular}}
\label{tab: arxiv gat transductive self-ensemble}
\end{center}
\end{table*}

\begin{table*}[htbp]
\caption{Reddit Test Accuracy, Ensemble vs Self-Ensembling, Transductive, GAT-2 Fanout 20}
\begin{center}\scalebox{.9}{ 
\begin{tabular}{|c|c|c|c|c|c|}
\hline
\textbf{Number}&\multicolumn{5}{|c|}{\textbf{Number of Models}} \\
\cline{2-6}\textbf{of Views} & 1 & 2 & 3 & 4 & 5 \\
\hline
1 & 96.26$\pm$0.08 & 96.48$\pm$0.11 & 96.53$\pm$0.04 & 96.56$\pm$0.06 & 96.57$\pm$0.06 \\\hline
2 & 96.46$\pm$0.05 & 96.65$\pm$0.11 & 96.71$\pm$0.06 & 96.76$\pm$0.04 & 96.75$\pm$0.02 \\\hline
3 & 96.53$\pm$0.07 & 96.71$\pm$0.11 & 96.74$\pm$0.08 & 96.81$\pm$0.04 & 96.81$\pm$0.03 \\\hline
4 & 96.60$\pm$0.04 & 96.73$\pm$0.06 & 96.80$\pm$0.04 & 96.82$\pm$0.04 & 96.84$\pm$0.03 \\\hline
5 & 96.59$\pm$0.07 & 96.74$\pm$0.06 & 96.79$\pm$0.04 & 96.83$\pm$0.04 & 96.85$\pm$0.03 \\\hline
\end{tabular}}
\label{tab: reddit gat transductive se}
\end{center}
\end{table*}

\begin{table*}[htbp]
\caption{Reddit Test Accuracy, Ensemble vs Self-Ensembling, Transductive, GCN-2 Fanout 20}
\begin{center}\scalebox{.9}{ 
\begin{tabular}{|c|c|c|c|c|c|}
\hline
\textbf{Number}&\multicolumn{5}{|c|}{\textbf{Number of Models}} \\
\cline{2-6}\textbf{of Views} & 1 & 2 & 3 & 4 & 5 \\
\hline
1 & 96.14$\pm$0.03 & 96.27$\pm$0.10 & 96.36$\pm$0.06 & 96.41$\pm$0.02 & 96.39$\pm$0.05 \\ \hline
2 & 96.31$\pm$0.05 & 96.44$\pm$0.06 & 96.53$\pm$0.03 & 96.52$\pm$0.08  & 96.53$\pm$0.02 \\ \hline
3 & 96.36$\pm$0.03 & 96.47$\pm$0.07 & 96.56$\pm$0.01 & 96.56$\pm$0.01 & 96.58$\pm$0.01 \\ \hline
4 & 96.37$\pm$0.04 & 96.50$\pm$0.01 & 96.55$\pm$0.02 & 96.61$\pm$0.03 &  96.61$\pm$0.03 \\ \hline
5 & 96.41$\pm$0.03 & 96.44$\pm$0.04 & 96.56$\pm$0.02 & 96.60$\pm$0.01 & 96.62$\pm$0.02 \\
\hline
\end{tabular}}
\label{tab: reddit gcn transductive se}
\end{center}
\end{table*}

\section{Additional Results: Consistency Training vs Single-View Single-Model}
 
\begin{figure*}
  \begin{subfigure}[t]{0.5\textwidth}
    \includegraphics[width=\textwidth]{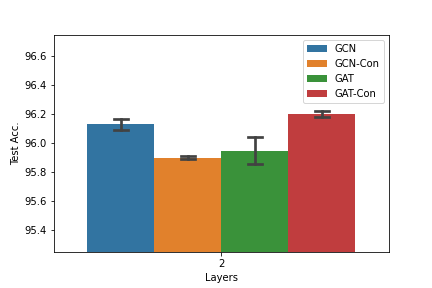}
    \caption{High label rate training.}
    \label{fig: reddit high label}
  \end{subfigure}\hfill
  \begin{subfigure}[t]{0.5\textwidth}
    \includegraphics[width=\textwidth]{figs/reddit_low_label_rate.png}
    \caption{Low label rate training.}
    \label{fig: reddit low label}
  \end{subfigure}\hfill
\caption{Inductive results on Reddit using fanout-20 models for a range of models and label rates. (a) Results with $66\%$ of labels used during training. (b) Results with $6.6\%$ of labels used for training. }
\label{fig: reddit consistency results multiple models}
\end{figure*}

In this Figure \ref{fig: reddit consistency results multiple models} we present additional results from our consistency training procedure on the reddit dataset. These results confirms results from the main body, in which consistency training performs best when label rates are low.

\bibliography{references.bib}